\pdfoutput=1
\documentclass[10pt,journal]{IEEEtran}
\usepackage[margin=0.75in]{geometry}
\usepackage{times}

\usepackage{amsmath,amssymb}
\usepackage{xcolor}
\usepackage{cite}
\usepackage{amsfonts}
\usepackage{algorithmic}
\usepackage{graphicx}
\usepackage{textcomp}
\usepackage{multirow}
\usepackage{mathrsfs}
\usepackage{booktabs}
\usepackage{array}
\usepackage{tabularx}
\usepackage{makecell}
\usepackage{ragged2e}
\usepackage{listings}
\usepackage{pgfplots}
\pgfplotsset{compat=1.17}

\usepackage{tikz}
\usetikzlibrary{arrows.meta, positioning, shapes.geometric}

\newcolumntype{Y}{>{\RaggedRight\arraybackslash}X}

\def\BibTeX{{\rm B\kern-.05em{\sc i\kern-.025em b}\kern-.08em
		T\kern-.1667em\lower.7ex\hbox{E}\kern-.125emX}}
\newcommand{\note}[1]{}

\begin{document}
	
	\title{Self-Supervised Speech Representations for Cross-Speaker Dysarthria Detection During Awake Craniotomy}
	\author{
Chinmayi~Kanthila,
Nassib~Abdallah,
Harrison~Misy,
Celine~Panheleux,\\
Vanessa~Saliou,
Romuald~Seizeur,
and Guillaume~Dardenne %
\thanks{Preprint, October 2026. This work was supported in part by
the University Hospital of Brest (CHU de Brest).}%
\thanks{C. Kanthila, N. Abdallah, H. Misy, V. Saliou,
R. Seizeur, and G. Dardenne are with LaTIM UMR1101,
INSERM, University of Western Brittany, Brest, France.}%
\thanks{N. Abdallah, C. Panheleux, V. Saliou, and R. Seizeur
are with the Neurosurgery Department, University Hospital of Brest,
Brest, France.}%
\thanks{Corresponding author: Chinmayi Kanthila
(e-mail: chinmayi.kanthila@inserm.fr).}%
}

	\maketitle
	

\begin{abstract}

    Detecting intra-operative speech impairment during awake craniotomy is essential for preserving language function. However, automated detection remains challenging because operating-room recordings contain substantial acoustic interference, clinically relevant speech events are rare, and available cohorts are small and heterogeneous across speakers. This study presents a systematic component-wise evaluation of a pipeline for distinguishing dysarthric from no-trouble speech in the DATABRASE corpus of awake-craniotomy recordings. The pipeline incorporates speaker diarization to isolate patient speech, a multi-view representation combining handcrafted acoustic descriptors with multilayer wav2vec~2.0 embeddings, speaker-conditional normalization and transferability-based feature selection to improve cross-speaker robustness, and a cascaded classifier comprising a gradient-boosted first stage and a neural second stage. Evaluation was conducted under strict speaker-independent conditions using leave-one-speaker-out cross-validation. The results show that cross-speaker performance is influenced more strongly by the speech representation than by classifier choice. The AUCs of three classifiers differed by no more than $4.7\%$, whereas replacing conventional acoustic descriptors with the multilayer self-supervised representation produced AUC improvements of $18.2\%$--$26.1\%$. Diarization-conditioned feature extraction and the proposed classifier cascade provided additional consistent gains. These findings indicate that reliable patient-specific speech isolation and strong pretrained representations are more important than increased classifier complexity in low-resource intra-operative settings. They also quantify the potential performance gains that may be achieved through patient-specific preoperative calibration.
\end{abstract}

	
\section{Introduction}
	
Awake craniotomy (AC) is the standard surgical approach for resecting tumors growing within or near eloquent cortex, where the surgeon must balance maximal resection against the preservation of motor functions including speech. Lesion size and proximity to cortical or subcortical language networks both raise the risk of postoperative cognitive decline, with direct consequences for functional independence and quality of life~\cite{martin-monzonLanguageMappingSystematic2022}.
Intra-operative monitoring is mainly dependent on direct electrical stimulation (DES) combined with active patient participation in speech and language tasks. DES can also elicit transient speech disturbances such as speech arrest, anomia, dysarthria, semantic and phonemic paraphasia, and comprehension
failures~\cite{colleeSpeechLanguageErrors2022}. Their clinical significance varies, as a transient dysarthric episode may reflect disruption of a distributed motor-speech network whose long-term consequences cannot be inferred from a naming-task score alone.

Detecting these events in real time still depends entirely on the surgical team. The neurosurgeon, neuropsychologist, and speech-language pathologist must run the language tests and assess the speech output simultaneously. The evaluation is subjective, has limited temporal resolution, and is prone to fatigue over a long procedure~\cite{gommersP0112BAnalysisSemispontaneous2022,keoughSURG22AwakeSpeech2025}.

In this context, the machine-learning framework offers the possibility of quantifying subtle variations in speech directly from intra-operative recordings, classifying emerging impairment patterns, and providing the surgical team with objective, decision-relevant feedback during resection. Rather than replacing expert clinical judgment, such a system could function as an assistive monitoring layer that enhances the sensitivity, consistency, and interpretability of intra-operative language assessment. Nevertheless, the intra-operative environment introduces a distinct set of constraints that conventional automatic speech recognition (ASR) frameworks do not adequately address. These include substantial acoustic noise from the operating-room environment and the patient's fluctuating neurological state, the absence of large-scale labeled intra-operative corpora, and the requirement for robust real-time performance under clinically critical conditions~\cite{bhatSpeechTechnologyAutomatic2025}. In fact, each of these constraints arises from a different stage of the processing chain. Noise and multi-speaker content relate to data quality, corpus scarcity relates to feature extraction and augmentation, and cross-patient robustness relates to classifier choice and evaluation.

The main contributions of this work are as follows:

\begin{itemize}

\item A pathology-preserving pitch augmentation using a HiFi-GAN vocoder that improves prosodic diversity while retaining dysarthric cues.

\item Patient-only feature extraction, using diarization to compute contamination-sensitive descriptors from patient frames alone and thereby remove clinician-speech confounds.

\item Two cascaded classifiers coupling a gradient-boosted first stage with a neural second stage through leakage-free inner cross-validation.

\end{itemize}
\section{Background and related work} 

Automated detection of dysarthric speech has developed almost entirely on chronic motor-speech disorders recorded under controlled conditions. Existing literature mainly focuses on what is methodologically achievable, including which acoustic properties carry the impairment, which representations expose them, and which evaluation protocols provide an honest estimate of performance on a new patient. However, there is a lack of studies that describe the conditions of an AC, where impairment is transient and stimulation-evoked, the recording contains clinicians as well as the patient, and the cohort is small and severely imbalanced. 

\subsection{Automated dysarthria detection in general conditions} 

Dysarthria is a group of motor-speech disorders arising from impaired control of the articulators. Most automated work uses the UA-Speech or TORGO corpora~\cite{kimDysarthricSpeechDatabase2008, rudziczTORGODatabaseAcoustic2012}, which contain speakers with chronic dysarthria of graded severity recorded under controlled conditions. 

Early systems described these recordings with interpretable acoustic descriptors. Prabhakera and Alku~\cite{prabhakeraDysarthricSpeechClassification2018} classified dysarthric speech using glottal-source parameters with a support vector machine, reporting 94.29\% under leave-one-speaker-out (LOSO) evaluation, and showing that source-level measures alone carry substantial diagnostic information. Prosodic and voice-quality descriptors have been used in the same way, selected for their correspondence to identifiable production mechanisms~\cite{kadiDiscriminativeProsodicFeatures2013, al-aliClassificationDysarthriaBased2023}. Millet and Zeghidour~\cite{milletLearningDetectDysarthria2019} instead learned the front end directly from the waveform, obtaining roughly a 10\% gain over fixed features and demonstrating that hand-designed descriptors do not exhaust the available signal, though at the cost of clinical interpretability. 

Self-supervised encoders have since become the dominant approach for speech feature extraction. Javanmardi et al.~\cite{javanmardiPretrainedModels2024} compared wav2vec~2.0 (W2V2) and HuBERT embeddings against conventional descriptors and found the pretrained representations consistently superior, while noting that the evaluation corpora remain small and English-only. Pasad et al.~\cite{pasadLayerWiseAnalysisSelfSupervised2021} showed by layer-wise probing that such encoders distribute information with depth, phonetic detail appearing in middle layers and speaker- and task-conditional information towards the output. This property is consequential for detection because the layers that encode intelligibility also encode speaker identity, so that a discriminative embedding may separate speakers rather than conditions. Kadirvelu et al.~\cite{kadirveluSpeakerIndependentDysarthria2025} addressed this directly. They combined a W2V2 encoder with a multi-task objective intended to group embeddings by severity rather than by speaker, and reported 70.5\% accuracy and 59.2\% F1 under LOSO evaluation. In this work, speaker invariance improved, but variability in severity remained unresolved.

Classifier design has followed a comparable progression, from classical statistical models to recurrent and convolutional architectures and, more recently, to transformer-based transfer learning~\cite{joshyAutomatedDysarthriaSeverity2022, sureshSpeakerIndependentDysarthria2025, venkatasivamanojTransformerbasedTransferLearning2024, mahumTranDSRHybridModel2024}. A separate line combines architectures rather than selecting among them, on the reasoning that one inductive bias suits a heterogeneous representation poorly. Shahin et al.~\cite{shahinNovelCascadedGaussian2020} placed a Gaussian mixture stage before a deep network for speaker identification under varying emotion, and the two-stage cascade outperformed either model used alone. A later study by the same group, comparing different shallow-deep pairings, found that the benefit came from coupling the two stages rather than from the particular models chosen~\cite{shahinNovelHybridDNN2021}. Ensemble formulations have recently been applied to dysarthria detection on the same principle~\cite{adsddStackedEnsemble2025}. In each case, the models are combined, but the representation they operate on is the same for both stages.

Across this literature, the evaluation protocol determines the reported result as much as the architecture. Tripathi et al.~\cite{tripathiImprovedSpeakerIndependent2020} report 97.4\% under speaker-dependent evaluation against 53.9\% under speaker-independent evaluation on UA-Speech, and Joshy and Rajan~\cite{joshyAutomatedDysarthriaSeverity2022} report 93.97\% against 49.22\% with a single architecture across UA-Speech and TORGO. Manoj et al.~\cite{venkatasivamanojTransformerbasedTransferLearning2024} report 99.0--99.4\% with transformer transfer learning, but without speaker-independent or external validation. Schu et al.~\cite{schuUsingUASpeechTORGO2023} showed that part of the performance reported on these corpora is attributable to non-speech and database-specific cues rather than to the impairment, and Roy et al.~\cite{royEnhancingSpeakerIndependent2025} found that accuracy under cross-corpus evaluation recovers only after adaptation to the target domain. Table~\ref{tab:dysarthria_review_summary} summarizes these studies.  These results are obtained on recordings that contain a single speaker, a stable impairment, and material elicited under controlled conditions.


\begin{table}[!t]
\centering
\caption{Recent dysarthric-speech classification studies. SI: speaker-independent; LOSO:
	leave-one-subject-out; SSL: self-supervised learning.}
\label{tab:dysarthria_review_summary}
\scriptsize
\setlength{\tabcolsep}{1.6pt}
\renewcommand{\arraystretch}{1.05}
\begin{tabularx}{\columnwidth}{@{}
		>{\raggedright\arraybackslash}p{1.10cm}
		>{\raggedright\arraybackslash}p{1.35cm}
		>{\raggedright\arraybackslash}p{1.25cm}
		>{\raggedright\arraybackslash}X
		@{}}
	\toprule
	\textbf{Ref.} & \textbf{Method} & \textbf{Result} & \textbf{Comment} \\
	\midrule
	
	\cite{tripathiImprovedSpeakerIndependent2020} &
	DeepSpeech + SVM &
	53.9\% SI; 97.4\% SD &
	Large SD--SI gap indicates strong speaker dependence; SD results may overestimate clinical generalisation. \\
	
	\cite{joshyAutomatedDysarthriaSeverity2022} &
	DNN/ CNN/ GRU/ LSTM &
	93.97\% SD; 49.22\% SI &
	Useful baseline study, but the same model drops sharply under SI testing, highlighting poor unseen-speaker transfer. \\
	
	\cite{kadirveluSpeakerIndependentDysarthria2025} &
	SSL transformer + MTL &
	70.5\% Acc.; 59.2\% F1 &
	LOSO evaluation is stronger than SD testing; results show improved speaker invariance but severity variability remains difficult. \\
	
	\cite{prabhakeraDysarthricSpeechClassification2018} &
	Glottal + SVM &
	94.29\% LOSO &
	Interpretable source-based features support clinical analysis; however, performance should be interpreted with dataset/task constraints. \\
	
	~\cite{venkatasivamanojTransformerbasedTransferLearning2024} &
	Transformer transfer learning &
	99.0--99.4\% SD &
	Near-ceiling SD accuracy is likely optimistic; without SI/LOSO or external validation, generalisability is limited. \\
	
	~\cite{javanmardiPretrainedModels2024} &
	W2V2, HuBERT, handcrafted &
	SSL $>$ baselines &
	Strong comparison of SSL and handcrafted features; confirms benefit of pretrained embeddings, but datasets remain small and English-only. \\
	
	~\cite{royEnhancingSpeakerIndependent2025} &
	DSSCNet; cross-corpus &
	56.8--64.2\%; 77.8--79.4\% tuned &
	The cross-corpus setting is deployment-relevant; the improvement after tuning suggests that domain mismatch remains a major limitation. \\
	
	~\cite{milletLearningDetectDysarthria2019} &
	Raw-speech end-to-end &
	$\sim$10\% gain &
	Early end-to-end evidence; learned front-ends improve over fixed features, but clinical interpretability is weaker. \\
	
	~\cite{schuUsingUASpeechTORGO2023} &
	UA/TG validation analysis &
	Artefact risk &
	Critical control study; shows that non-speech or speaker/database cues can inflate reported performance on UA and TG. \\
	
	\bottomrule
\end{tabularx}
\end{table}

\subsection{Speech impairment detection during awake craniotomy} 

Speech assessment during AC has been addressed primarily as functional mapping rather than signal-level classification~\cite{colleeSpeechLanguageErrors2022, martin-monzonLanguageMappingSystematic2022}, and only two studies have approached the classification problem directly. Nishimura et al.~\cite{nishimuraClassificationPatientsReaction2014} trained support vector and relevance vector machines on MFCC features from five awake procedures to detect speech arrest, establishing feasibility but without speaker-independent evaluation. Maoudj et al.~\cite{maoudjClassificationSpeechArrests2025} extended this work with a W2V2 classifier on 1{,}883 three-second clips from 23 procedures, including a cross-lingual Japanese--French evaluation. In-domain accuracy was strong but degraded under cross-lingual transfer, indicating that the model had learned regularities specific to the training condition rather than the impairment itself. 

Both studies formulate the task as abnormal versus normal speech, grouping heterogeneous deficits such as dysarthria, anomia, paraphasia, and perseveration~\cite{duffauAwakeSurgeryIncidental2012}. A correct prediction therefore indicates only that speech has become abnormal, without identifying which deficit occurred, although these impairments reflect different functional networks and are distinguished clinically during stimulation. The present work accordingly restricts the target to dysarthria, so that the predicted label corresponds to a single motor-speech phenomenon rather than to a heterogeneous abnormality class.

The intra-operative setting also breaks assumptions that the corpus literature relies on. Recordings are made in an active operating room, so the patient's voice is accompanied by clinician speech, equipment noise, and silence. Clinician speech is not incidental, since the examiner prompts, repeats, and follows up in response to the patient, and its presence is therefore correlated with the condition being detected. Isolating the patient is a diarization problem, and diarization is an established preliminary step in clinical speech analysis, used for cognitive assessment recordings before any diagnostic model is applied~\cite{xuSpeakerTurnAwareDiarization2024}. Its reliability is not assured for this population, as general-purpose systems degrade on conversations involving speakers with speech disorders~\cite{riadComparisonSpeakerDiarization2022}, and the labels it produces are anonymous, so identifying which track belongs to the patient requires a further decision~\cite{roleIdentificationClinical}. In each of these applications the segmentation is retained, and the description of the exchange that produced it is discarded. 

Cohort size is a second constraint. AC procedure has few patients, impaired speech occupies a small fraction of each recording, and the resulting class imbalance is severe. Augmentation is the standard response to scarcity, but it is constrained here in a way it is not for ordinary speech. The label of a clip is defined by the acoustic consequences of impairment, so any transformation that improves class balance while altering those consequences removes the property the classifier is meant to detect. Voice-conversion methods, which modify speaker characteristics directly, are unsuitable for this reason. What is required is not more data but data whose label is preserved by construction.

These constraints explain why reliable automated dysarthria classification during AC remains largely unexplored. The two prior studies showed that the detection is feasible. However, gaps still exist in the literature concerning the recording conditions, the size of the available cohorts, and the design of the classifier. The recording contains clinician speech that is correlated with the label and must be separated from the patient's own. The cohort is too small and too imbalanced to train on directly, and the augmentation required to address this must preserve the acoustic evidence of impairment. The resulting representation combines descriptors of different origin and reliability, which no single classifier is well suited to exploit.

\section{Materials and methods}
\subsection{Dataset}

The French dataset used for dysarthric classification, DATABRASE, was collected during AC and comprises intra-operative speech recordings from 23 patients undergoing AC at the University Hospital (CHU) of Brest, France. For each procedure, the continuous recording was segmented into 3-second clips. This duration was chosen to capture the patient's response to a single DES event together with any associated speech disturbance, short enough to localize transient impairment, yet long enough to support stable acoustic feature extraction. Although intra-operative DES can elicit a range of language disturbances, including anomia, paraphasia, and perseveration, the present work restricts the classification target to DYS versus NT speech.

\begin{table}[t]
	\centering
	\caption{Speaker-wise distribution of the data.}
	\label{tab:databrase_distribution}
	\begin{tabular}{lcccc}
		\hline
		\textbf{Speaker ID} & \textbf{Total clips} & \textbf{DYS} & \textbf{NT} & \textbf{DYS \%} \\
		\hline
		Patient 1  & 796 & 38  & 758 & 4.8  \\
		Patient 2  & 355 & 31  & 324 & 8.7  \\
		Patient 3  & 112 & 11  & 101 & 9.8  \\
		Patient 4  & 254 & 39  & 215 & 15.4 \\
		Patient 5  & 173 & 28  & 145 & 16.2 \\
		Patient 6  & 35  & 6   & 29  & 17.1 \\
		Patient 7  & 607 & 117 & 490 & 19.3 \\
		Patient 8  & 665 & 142 & 523 & 21.4 \\
		Patient 9  & 244 & 78  & 166 & 32.0 \\
		Patient 10 & 773 & 265 & 508 & 34.3 \\
		Patient 11 & 267 & 130 & 137 & 48.7 \\
		\hline
		\textbf{Total} & \textbf{4{,}281} & \textbf{885} & \textbf{3{,}396} & \textbf{20.7} \\
		\hline
	\end{tabular}
\end{table}

Each clip was annotated independently by the neurosurgeon using Audacity and subsequently reviewed by two additional speech experts. Patients with no DYS segments were removed, since they provided no positive samples. These patient-level exclusion criteria were then applied to support robust speaker-independent modeling. The retained cohort comprised speakers whose dysarthric proportions ranged from approximately 5\% to 49\%, with the lowest-proportion retained speaker additionally excluded from within-fold validation use, since the near-absent positive class could not provide a meaningful signal for threshold or model selection. The resulting cohort, summarized in Table~\ref{tab:databrase_distribution}, comprised 11 speakers and 4{,}281 raw 3-second clips.

The classification framework comprises six stages, from raw-audio conditioning to speaker-grouped evaluation. HiFi-GAN pitch augmentation expands the training folds with label-preserving synthetic clips. These clips add prosodic variability while retaining dysarthric cues. Speaker diarization then separates patient speech from clinician speech. It isolates the frames on which contamination-sensitive features are measured, and yields conversational descriptors of the clinician–patient exchange. A quality filter discards clips with insufficient patient occupancy. The retained clips are described by complementary feature families: conventional acoustic descriptors, conversational features from diarization, and multi-layer self-supervised embeddings. These features are refined by per-fold preprocessing. Each patient is normalized against their own baseline, and only features that transfer across speakers are retained. Classification uses three single-stage baselines and two proposed cascades, each coupling a gradient-boosted stage with a neural one. All stages beyond feature extraction run within each cross-validation fold. Held-out speakers therefore influence neither the preprocessing statistics nor the model hyperparameters. Each stage is introduced cumulatively, so its individual effect can be isolated.


\begin{figure*}[t]
	\centering
	\includegraphics[width=0.95\textwidth]{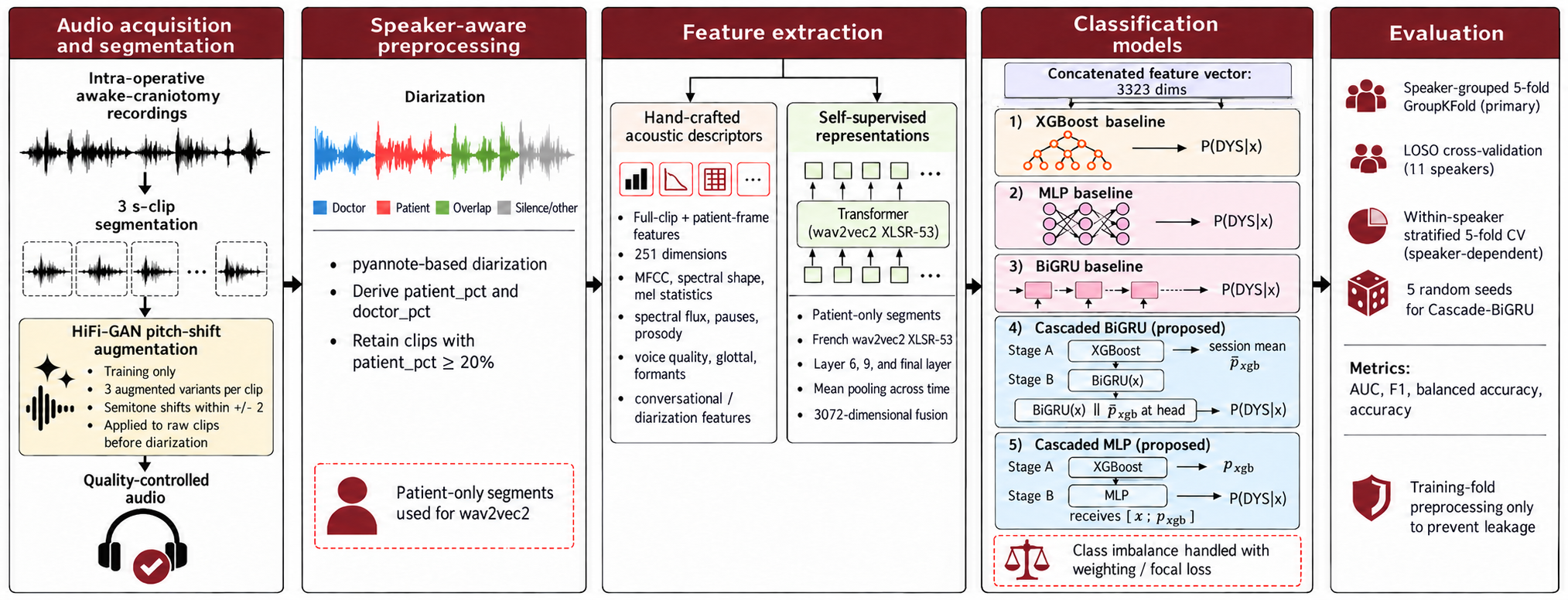}
	\caption{Overall classification framework, comprising the data
		quality, feature extraction, augmentation, and classification stages.}
	\label{fig:methodology}
\end{figure*}

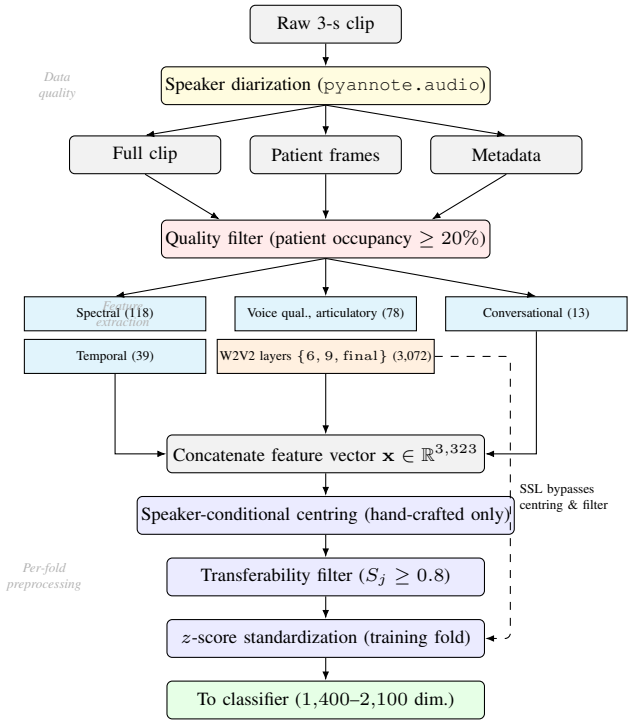
\begin{figure}[t]
	\centering
	\begin{tikzpicture}[
		font=\scriptsize,
		node distance=3mm and 3mm,
		input/.style={
			rectangle, draw, rounded corners=2pt,
			minimum height=5mm, minimum width=20mm,
			align=center, inner sep=1.5pt, fill=gray!10
		},
		stage/.style={
			rectangle, draw, rounded corners=2pt,
			minimum height=5mm,
			align=center, inner sep=1.5pt
		},
		filt/.style={
			rectangle, draw, rounded corners=2pt,
			minimum height=5mm, minimum width=42mm,
			align=center, inner sep=1.5pt, fill=red!8
		},
		feat/.style={
			rectangle, draw,
			minimum height=4.5mm,
			align=center, inner sep=1pt, font=\tiny
		},
		ssl/.style={
			rectangle, draw,
			minimum height=4.5mm,
			align=center, inner sep=1pt, font=\tiny, fill=orange!12
		},
		proc/.style={
			rectangle, draw, rounded corners=2pt,
			minimum height=5mm,
			align=center, inner sep=1.5pt, fill=blue!8
		},
		output/.style={
			rectangle, draw, rounded corners=2pt,
			minimum height=5mm,
			align=center, inner sep=1.5pt, fill=green!10
		},
		stagelbl/.style={font=\tiny\itshape, text=gray!60, align=center},
		arrow/.style={-{Latex[length=1.2mm]}, thin},
		arrowt/.style={-{Latex[length=1.2mm]}, thin, dashed},
		]
		
		\node[input] (raw) {Raw 3-s clip};
		
		\node[stage, below=of raw, fill=yellow!15, minimum width=42mm]
		(diar) {Speaker diarization (\texttt{pyannote.audio})};
		\draw[arrow] (raw) -- (diar);
		
		\node[input, below left=4mm and -8mm of diar] (full) {Full clip};
		\node[input, below=4mm of diar] (pat) {Patient frames};
		\node[input, below right=4mm and -8mm of diar] (meta) {Metadata};
		
		\draw[arrow] (diar.south) -- (full.north);
		\draw[arrow] (diar.south) -- (pat.north);
		\draw[arrow] (diar.south) -- (meta.north);
		
		\node[filt, below=6mm of pat] (qf)
		{Quality filter (patient occupancy $\geq 20\%$)};
		\draw[arrow] (full.south) -- ([xshift=-14mm]qf.north);
		\draw[arrow] (pat.south)  -- (qf.north);
		\draw[arrow] (meta.south) -- ([xshift=14mm]qf.north);
		
		\node[feat, below left=5mm and -6mm of qf, fill=cyan!10, minimum width=24mm]
		(spec) {Spectral (118)};
		\node[feat, below=1mm of spec, fill=cyan!10, minimum width=24mm]
		(temp) {Temporal (39)};
		
		\node[feat, below=5mm of qf, fill=cyan!10, minimum width=24mm]
		(vqa) {Voice qual., articulatory (78)};
		\node[ssl, below=1mm of vqa, minimum width=24mm]
		(sslf) {W2V2 layers $\{6, 9, \mathrm{final}\}$ (3{,}072)};
		
		\node[feat, below right=5mm and -6mm of qf, fill=cyan!10, minimum width=24mm]
		(conv) {Conversational (13)};
		
		\draw[arrow] (qf.south) -- (spec.north);
		\draw[arrow] (qf.south) -- (vqa.north);
		\draw[arrow] (qf.south) -- (conv.north);
		
		\node[stage, below=8mm of sslf, minimum width=42mm, fill=gray!10]
		(concat) {Concatenate feature vector $\mathbf{x} \in \mathbb{R}^{3{,}323}$};
		\draw[arrow] (temp.south) |- (concat.west);
		\draw[arrow] (sslf.south) -- (concat.north);
		\draw[arrow] (conv.south) |- (concat.east);
		
		\node[proc, below=of concat, minimum width=42mm]
		(cen) {Speaker-conditional centring (hand-crafted only)};
		\draw[arrow] (concat) -- (cen);
		
		\node[proc, below=of cen, minimum width=42mm]
		(tf) {Transferability filter ($S_j \geq 0.8$)};
		\draw[arrow] (cen) -- (tf);
		
		\node[proc, below=of tf, minimum width=42mm]
		(zsc) {$z$-score standardization (training fold)};
		\draw[arrow] (tf) -- (zsc);
		
		\node[output, below=of zsc, minimum width=42mm]
		(clf) {To classifier ($1{,}400$--$2{,}100$ dim.)};
		\draw[arrow] (zsc) -- (clf);
		
		\draw[arrowt, rounded corners=2pt]
		(sslf.east) -- ++(10mm,0) |- (zsc.east)
		node[pos=0.25, right, font=\tiny, align=left]
		{SSL bypasses\\centring \& filter};
		
		\node[stagelbl, left=10mm of diar] {Data\\quality};
		\node[stagelbl, left=10mm of vqa] {Feature\\extraction};
		\node[stagelbl, left=10mm of tf] {Per-fold\\preprocessing};
		
	\end{tikzpicture}
	\caption{End-to-end data preprocessing and feature-extraction framework.}
	\label{fig:pipeline}
\end{figure}

\subsection{Data augmentation}

Dysarthric clips are the minority class throughout the cohort. Across the retained speakers, 885 of 4{,}281 clips are dysarthric, a ratio of 3.8:1 (Table~\ref{tab:databrase_distribution}). The imbalance is also uneven across patients, ranging from 1.1:1 to nearly 20:1. Training on these clips alone therefore risks a classifier biased toward the majority class and fitted to the few speakers that contribute most positive samples. To mitigate this, the training folds are expanded by pitch augmentation using the HiFi-GAN neural vocoder~\cite{kongHiFiGANGenerativeAdversarial2020}. Unlike conversion-based approaches such as CycleGAN and StarGAN-VC, which can distort pathology-bearing cues, HiFi-GAN resynthesizes along a single perceptual axis~\cite{qinHiFiGANMethodbasedAlgorithm2022}. This retains lexical content, response timing, and the dysarthric acoustic structure the classifier must learn. The V2 variant is used, fine-tuned on the training split.  

Given an input mel-spectrogram $s$, a controlled pitch shift in the conditioning domain produces a modified representation $s'$. The waveform is then resynthesized as $G(s')$. Pitch offsets are drawn uniformly from $\pm 2$ semitones. This range stays clinically plausible for AC while remaining label-preserving~\cite{qianDataAugmentationUsing2019b}. Three pitch-shifted variants are generated per source clip. Each variant inherits the speaker identity of its source and is bound to the same fold under the GroupKFold partition, which prevents train--test leakage. Augmentation is applied to the raw clips, so both the original and augmented clips then pass through diarization and feature extraction.

\subsection{Diarization and feature extraction}

\subsubsection{Speaker separation and clip selection}

Operating-room audio rarely contains the patient's voice in isolation. Clinician speech is present in most clips, and its share of each recording varies with the clinical situation. This poses two distinct problems. First, dysarthria is a disorder of motor speech production, so its acoustic markers describe the patient's articulatory and phonatory behavior. Measured over a mixture of speakers, descriptors such as jitter, shimmer, formant position, and glottal-source behavior no longer characterize the patient at all. Second, a classifier trained on unsegmented audio may separate the classes using recording context rather than patient speech.  

Speaker diarization addresses both. Each three-second clip is processed by \texttt{pyannote.audio}~\cite{bredinPyannoteAudioNeuralBuilding2020}, which yields the full waveform, the frames attributed to the patient, and turn-level descriptors of the clinician--patient exchange. These outputs are used in two ways. The patient frames define the region on which contamination-sensitive features are computed, so that each descriptor is measured where it is physiologically meaningful. The turn-level descriptors are retained as features in their own right, since the clinician's conduct during AC is reactive. The prompts are repeated, rephrased, or followed up when a response appears impaired, so response latency and follow-up behavior carry information about the patient's state. Diarization therefore serves both as a segmentation step preceding feature extraction and as a source of features, rather than as a preprocessing operation whose output is discarded.  

The diarization output also supports a clip-level quality criterion. The attribute \texttt{patient\_pct} gives the proportion of clip duration attributed to the patient. Clips with \texttt{patient\_pct} $< 20\%$ were excluded, since at low patient occupancy the acoustic content is dominated by clinician speech and pathology-relevant cues cannot be measured reliably. The criterion retained 2{,}706 of 4{,}281 clips (63.2\%), with a post-filter dysarthric prevalence of 22.1\% (597 DYS, 2{,}109 NT). The excluded 36.8\% were held out for sensitivity analysis and were not used for training.

\subsubsection{Feature representation}  

Each clip is represented by three groups of features, organised by the region on which they can be measured validly.  

The first group covers spectral and temporal properties of the clip as a whole. These include cepstral coefficients, spectral-shape descriptors, mel-band statistics, energy and rhythm measures, and pause statistics. Together they characterize vocal-tract shaping, loudness, and speech timing, all commonly altered in dysarthria~\cite{remyaArtificialIntelligenceSpeech2025}. Cepstral coefficients are the standard baseline in dysarthric speech analysis and serve that role here~\cite{bhatSpeechTechnologyAutomatic2025, qianSurveyAutomaticSpeech2023, joshyAutomatedDysarthriaSeverity2022, javanmardiPretrainedModels2024}. These descriptors remain interpretable under brief non-patient speech, since they summarize the acoustic scene rather than one talker's production, and are therefore computed on the full clip (Table~\ref{tab:feature_inventory}).

\begin{table}[t]
	\centering
	\caption{Feature inventory, grouped by the region on which each descriptor is
		computed: the full 3-s clip, the patient frames, or the diarization output.}
	\label{tab:feature_inventory}
	\scriptsize
	\setlength{\tabcolsep}{3pt}
	\renewcommand{\arraystretch}{1.05}
	\begin{tabularx}{\linewidth}{@{}l r X@{}}
		\toprule
		\textbf{Descriptor} & \textbf{Dim.} & \textbf{Details} \\
		\midrule
		\multicolumn{3}{@{}l}{\textit{Full-clip descriptors}} \\
		MFCC ($+\Delta,\Delta\Delta$) &   65 & Cepstral baseline \\
		Spectral shape                &   30 & Centroid, roll-off, bandwidth \\
		Mel-spectrogram               &   23 & Mel-band statistics \\
		Temporal (3\,s)               &   24 & RMS, ZCR, duration \\
		Spectral flux                 &    6 & Frame-to-frame change \\
		Pause                         &    9 & Pause and rhythm statistics \\
		\midrule
		\multicolumn{3}{@{}l}{\textit{Patient-frame descriptors}} \\
		Prosody                       &   18 & F0, intensity contour \\
		Voice quality                 &   12 & Jitter, shimmer, HNR \\
		Formants                      &   22 & F1--F3, vowel space \\
		Glottal                       &   26 & GNE, NAQ, QOQ, CPPS \\
		W2V2 (FR) layer 6             & 1{,}024 & Phonetic detail \\
		W2V2 (FR) layer 9             & 1{,}024 & Phonological structure \\
		W2V2 (FR) final layer         & 1{,}024 & Intelligibility-related \\
		\midrule
		\multicolumn{3}{@{}l}{\textit{Conversational descriptors}} \\
		Diarization                   &   13 & Turn-taking, latency, overlap \\
		\bottomrule
	\end{tabularx}
\end{table}

The second group covers source- and articulator-level properties, which describe the patient's production directly and are consequently computed on patient frames only. It comprises prosodic measures of fundamental frequency and intensity, voice-quality measures of jitter, shimmer, and harmonics-to-noise ratio~\cite{al-aliClassificationDysarthriaBased2023, joshyAutomatedDysarthriaSeverity2022}, formant positions and vowel-space area, and glottal-source descriptors obtained from inverse-filtered glottal flow with closure instants supplied by the speech processing tool, REAPER. This group also includes the self-supervised representation described below, which is likewise sensitive to who is speaking (Table~\ref{tab:feature_inventory}).  

The third group is derived from the diarization output rather than from the audio. It encodes the dynamics of the intra-operative exchange: response latency between clinician prompt and patient response, clinician follow-up rate, turn-taking statistics, and overlap proportions (Table~\ref{tab:feature_inventory}). As noted above, these quantities reflect the clinician's real-time assessment of the patient's response, and they are available only because diarization is part of the framework.

\subsubsection{Self-supervised representation}  

Conventional descriptors encode properties chosen in advance, and they capture dysarthric production only to the extent that those choices are correct. A self-supervised encoder instead learns speech structure from large external corpora, and can represent regularities that no explicit descriptor makes available. A frozen French-pretrained W2V2 encoder~\cite{conneau2021unsupervised, facebook_wav2vec2_large_xlsr_53_french} is therefore applied to the patient frames of each clip.  

Depth determines what such an encoder represents. Layer-wise probing shows that phonetic detail emerges in mid-network layers, phonological and lexical structure in deeper layers, and task- and speaker-conditional information in the final layers~\cite{pasadLayerWiseAnalysisSelfSupervised2021}. Dysarthria disturbs articulatory precision, phonological realisation, and intelligibility together, so no single layer carries the full profile. Following the aggregation strategy of~\cite{angraExploringAggregatedWav2vec2025}, layers 6, 9, and the final layer are retained. Frame-level activations of each retained layer $\ell$ are mean-pooled over time to $\mathbf{z}_\ell \in \mathbb{R}^{1024}$, and the three are concatenated, \begin{equation} 	
\mathbf{z}_{\mathrm{SSL}} = \mathbf{z}_6 \,\Vert\, \mathbf{z}_9 \,\Vert\, 	\mathbf{z}_{\mathrm{final}}, 
\end{equation} 

\noindent giving the complete representation $\mathbf{x} = \mathbf{x}_{\mathrm{HC}} \,\Vert\, \mathbf{z}_{\mathrm{SSL}}$.  

The encoder is kept frozen. With 597 dysarthric clips from eleven speakers, fine-tuning a 95M-parameter encoder would risk fitting speaker identity rather than dysarthric structure. The same sensitivity motivates the further preprocessing since the embedding is discriminative but encodes who is speaking alongside how they are speaking.

\subsubsection{Per-fold preprocessing}  

Both feature groups carry speaker identity alongside dysarthric structure. Three operations reduce this before classification, all estimated within the training fold only. Speaker-conditional centering expresses each conventional descriptor relative to the patient's own baseline. For each training-fold speaker $s$, a per-feature mean $\mu_s$ is estimated over that speaker's NT clips and subtracted, 
\begin{equation} 	
\tilde{x}_{\mathrm{HC},i} = x_{\mathrm{HC},i} - \mu_{s(i)}. 	\label{eq:centring} \end{equation} 

This follows the clinical logic of the task. Impairment is judged against how a given patient normally speaks, not against a population mean, and AC provides the pre-operative recording such a baseline requires.  The transferability filter then removes descriptors that do not generalize across patients. A feature is retained only if the sign of its within-speaker Spearman correlation with the label agrees across at least 80\% of training speakers, 
\begin{equation} 	
S_j = \frac{1}{|S_{\text{train}}|}\sum_{s\in S_{\text{train}}} 	\mathbb{1}\!\left[\operatorname{sign}(\rho_{j,s}) = 	\operatorname{sign}(\bar\rho_j)\right],\quad S_j \geq 0.8. 	\label{eq:transfer} 
\end{equation} 

A descriptor that rises with impairment in one patient and falls in another is uninformative for an unseen patient, however well it separates the classes within a speaker. The criterion typically retains 80 to 150 of the 251 conventional descriptors.  
Centring and the transferability filter are applied to the conventional descriptors only. Both require a per-speaker estimate over NT clips, and the self-supervised embedding is high-dimensional and opaque enough that per-feature sign consistency is not a meaningful criterion. The embedding is instead standardized with the rest of the representation. Standardization rescales every feature to zero mean and unit variance using training-fold statistics, which are then applied unchanged to the validation and test folds.  

All three operations are estimated inside the cross-validation loop. No statistic is computed from a held-out speaker, so the evaluation reflects performance on a patient the framework has not seen.

\subsection{Classification models} 
\label{sec:models}  

Three single-stage classifiers of distinct inductive bias are evaluated, together with two proposed cascades. The representation defined above is high-dimensional and heterogeneous. It mixes frequency-domain descriptors, temporal-domain measures, conversational descriptors, and a large self-supervised block, in which many dimensions are weakly informative. No single inductive bias suits all of it. Gradient-boosted trees are robust to uninformative dimensions but partition the space along individual axes. Neural models represent smooth dependencies among features, but are sensitive to noise when the sample is small. The cascades are designed to combine these behaviors rather than choose between them. All models are trained and evaluated under identical speaker-grouped protocols, so that architecture is the only variable.  

\subsubsection{XGBoost baseline}  

XGBoost~\cite{chenXGBoostScalableTree2016} serves as a baseline and as the first stage of both cascades. It is trained with the \texttt{reg:logistic} objective, with the positive-class weight set from the training-fold class ratio. Its splits are invariant to the differing numeric ranges of the conventional and self-supervised blocks, and its directional default splits tolerate sparse inputs. Boosted trees also remain competitive with deep architectures on small tabular problems~\cite{shwartz-zivTabularDataDeep2022}, which matches the regime here. Hyperparameters (\texttt{max\_depth}, \texttt{eta}, \texttt{reg\_lambda}, \texttt{subsample}, \texttt{colsample\_bytree}, \texttt{min\_child\_weight}) were selected by Bayesian search with a Tree-structured Parzen Estimator sampler~\cite{akiba2019optuna}. The search ran independently within each fold on the validation speaker only, so the test speaker never influenced selection. Thirty trials were used per fold, since pilot runs showed a clear AUC gain from 5 to 15 trials and below $0.005$ from 15 to 50. Each trial allowed up to 200 boosting rounds with early stopping on validation AUC.  

\subsubsection{Multilayer perceptron}  

The MLP provides a non-recurrent neural counterpart to the tree baseline. It operates on the full concatenated vector and represents smooth interactions among features that axis-aligned splits cannot express. This capacity carries a cost: with few dysarthric clips, the network is more exposed to noisy dimensions than the tree. Each hidden layer is followed by batch normalisation, ReLU, and dropout, and the width is scaled to the input dimensionality. The output head is trained with focal binary cross-entropy, 
\begin{equation} 	
\mathcal{L}_{\mathrm{focal}}(p, y) 	
= -\alpha_{y}\,(1 - p_{t})^{\gamma}\,\log p_{t}, 	
\quad 	
p_{t} = \begin{cases} p & y = 1 \\ 1 - p & y = 0, \end{cases} 	
\label{eq:focal_loss} 
\end{equation} 

\noindent with $\gamma = 2.0$, $\alpha_{1}$ derived from the training-fold class ratio, and $\alpha_{0} = 1$~\cite{linFocalLossDense2017}. The focal term down-weights easy majority-class clips, which is appropriate given the imbalance reported in Table~\ref{tab:databrase_distribution}. Hyperparameters were tuned under the same protocol as the other models (Table~\ref{tab:hparams}), with Adam and early stopping on validation AUC.  

\subsubsection{Bidirectional GRU}  

The BiGRU is the recurrent baseline. The other two models consume clip-level summaries, in which the order of events within a clip has already been pooled away. A dysarthric episode is a transient event with onset and release. Each three-second clip is therefore divided into overlapping sub-segments, and the resulting sequence is passed through a two-layer bidirectional encoder, so that within-clip temporal structure remains available to the model. Sequence representations are mean-pooled and projected through a linear head with sigmoid activation. Training uses the same focal objective as the MLP, with the positive-class weight set from the training-fold class ratio, Adam optimization, and early stopping on validation AUC.  

\subsubsection{Cascade-MLP} 
\label{sec:cascade}  

The first cascade places an XGBoost stage before an MLP stage. The two models fail in different ways on this representation. XGBoost is robust to the many weakly informative dimensions of the self-supervised block, but cannot express smooth dependencies among features. The MLP can express them, but must locate them in a 3{,}323-dimensional space from few positive examples, and tends to fit speaker identity instead. The cascade removes part of that burden. The tree reduces the representation to a single calibrated estimate of dysarthria, and the neural stage receives that estimate alongside the original features, 
\begin{equation} 	
\mathbf{x}_{\mathrm{cas}} = \mathbf{x} \,\Vert\, p_{\mathrm{XGB}}, 	
\label{eq:cascade_input} 
\end{equation}
\noindent where $p_{\mathrm{XGB}} = f_{\mathrm{XGB}}(\mathbf{x})$ and $p_{\mathrm{cas}} = f_{\mathrm{MLP}}(\mathbf{x}_{\mathrm{cas}})$. The second stage is therefore not required to rediscover the decision boundary from the raw input, and can instead refine a summary that is already stable across speakers.  

Two design choices keep the construction valid under speaker-independent evaluation. First, each training clip's $p_{\mathrm{XGB}}$ is produced by an inner LOSO procedure within the training fold. A tree evaluated on its own training clips would return overconfident probabilities, and the second stage would learn to trust an input whose character it never sees again at test time. The inner procedure makes training-time and test-time probabilities statistically comparable. Second, an auxiliary regression head $g(\mathbf{h}_2)\in\mathbb{R}^4$ on the second hidden layer is trained jointly with the classification head, 
\begin{equation} 	
\mathcal{L}_{\mathrm{cas}} 	
= \mathcal{L}_{\mathrm{focal}}(p_{\mathrm{cas}}, y) 	
+ \alpha\,\mathcal{L}_{\mathrm{MSE}}(g(\mathbf{h}_2), \mathbf{c}), 	
\label{eq:cascade_loss} 
\end{equation} 
\noindent with $\mathbf{c} = (0.5, 0.5, 0.5, 0.5)^\top$ a soft uniform prior and $\alpha = 0.1$. The auxiliary term discourages the hidden layer from collapsing onto sharp class indicators, which in a small cohort are often speaker-specific.  

\subsubsection{Cascade-BiGRU} 
\label{sec:cascade_bigru} 

The second cascade replaces the feed-forward second stage with a recurrent one, so that the first-stage estimate is combined with a sequence model rather than a static classifier. The first stage and its inner LOSO construction are unchanged. 

Each session $k$ is represented by an ordered sequence of segment feature vectors $\mathbf{X}^{(k)} = (\mathbf{x}^{(k)}_1, \dots, \mathbf{x}^{(k)}_{T_k})$, where $T_k$ is the number of retained segments. The bidirectional encoder maps this sequence to hidden states, which are mean-pooled into a single session representation, 
\begin{equation} 
\mathbf{h}^{(k)} = \frac{1}{T_k}\sum_{t=1}^{T_k} \operatorname{BiGRU}\!\left(\mathbf{X}^{(k)}\right)_t . 
\label{eq:bigru_pool} 
\end{equation} 

The first-stage estimate enters as a session-level summary, obtained by averaging the segment probabilities of that session, 
\begin{equation} 
\bar{p}^{(k)}_{\mathrm{XGB}} = \frac{1}{T_k}\sum_{t=1}^{T_k} f_{\mathrm{XGB}}\!\left(\mathbf{x}^{(k)}_t\right). 
\label{eq:xgb_session} 
\end{equation} 

The pooled representation and the summary are concatenated at the classification head, which produces the session logit, 
\begin{equation} 
p^{(k)}_{\mathrm{cas}} = \sigma\!\left( \mathbf{w}^\top \left[\mathbf{h}^{(k)} \,\Vert\, \bar{p}^{(k)}_{\mathrm{XGB}}\right] + b \right). 
\label{eq:cascade_bigru_head} 
\end{equation} 

The placement of $\bar{p}^{(k)}_{\mathrm{XGB}}$ is a deliberate choice. The alternative is to append the segment probability to every element of the input sequence, replacing $\mathbf{x}^{(k)}_t$ by $[\mathbf{x}^{(k)}_t \Vert p^{(k)}_t]$ in \eqref{eq:bigru_pool}. Both placements were implemented, and the variant of \eqref{eq:cascade_bigru_head} is retained. A scalar repeated across all timesteps is easily exploited by a recurrent encoder and would compete with the temporal structure the encoder exists to model. Admitting the estimate only at the decision boundary keeps the encoder operating on the feature stream, and confines the first-stage contribution to the point where the two sources are combined. 

The head is trained with the same focal objective as the single-stage BiGRU, with the positive-class weight set from the training-fold class ratio. Tuning used the same inner speaker-independent protocol, with 30 trials per fold under three-fold inner cross-validation. The retained configuration was evaluated over five random seeds, with predicted probabilities averaged before thresholding to match the Cascade-MLP.

\begin{table}[t]
	\centering
	\caption{Hyperparameter search spaces and tuning settings.}
	\label{tab:hparams}
	\footnotesize
	\begin{tabular}{lll}
		\toprule
		\textbf{Model} & \textbf{Hyperparameter} & \textbf{Search space} \\
		\midrule
		\multirow{7}{*}{\shortstack[l]{MLP /\\Cascade-MLP}}
		& Hidden layers            & $\{2, 3, 4\}$ \\
		& Base hidden dim.         & $\{64, 128, 256, 512\}$ \\
		& Dropout                  & $[0.15, 0.55]$ \\
		& Learning rate            & log-U $[10^{-4}, 5\!\times\!10^{-3}]$ \\
		& Weight decay             & log-U $[10^{-6}, 10^{-3}]$ \\
		& Focal-loss $\gamma$      & $[1.0, 3.0]$ \\
		& Auxiliary weight $\alpha$ & $[0.0, 0.3]$ \\
		\midrule
		\multirow{7}{*}{\shortstack[l]{BiGRU /\\Cascade-BiGRU}}
		& Hidden dimension     & $\{32, 64, 128, 256\}$ \\
		& Recurrent layers     & $\{1, 2\}$ \\
		& Max sequence length  & $\{10, 15, 20\}$ segments \\
		& Dropout              & $[0.15, 0.55]$ \\
		& Focal-loss $\gamma$  & $[1.0, 3.0]$ \\
		& Learning rate        & log-U $[10^{-4}, 5\!\times\!10^{-3}]$ \\
		& Weight decay         & log-U $[10^{-6}, 10^{-3}]$ \\
		\midrule
		\multicolumn{3}{l}{Batch size $\{16, 32, 64\}$ for all neural models.} \\
		\bottomrule
	\end{tabular}
\end{table}

\subsection{Evaluation protocol and metrics} 
\label{sec:cv} 

Performance is reported with four metrics. AUC is the primary metric. It is threshold-independent and measures ranking quality across all operating points, which suits a task with marked class imbalance and a speaker-adaptive decision threshold. The F1-score, balanced accuracy (BAcc), and overall accuracy (Acc) are reported as secondary, threshold-dependent metrics, computed at that operating point. Because dysarthric clips are the minority class, accuracy alone is unreliable, and BAcc and F1 are included to reflect performance on the positive class. 

Three evaluation protocols are used, and they answer different questions. Speaker-independent (SID) evaluation measures generalization to unseen patients and is reported in two variants that differ in how many speakers are held out at a time. The first, referred to as 5-fold or SID, is the primary protocol. A 5-fold GroupKFold partition was built at the speaker level and stratified by per-speaker dysarthric proportion, so that no clip from a given speaker appeared in more than one of the training, validation, and test sets of a fold. Each fold contained two to three test speakers. One remaining training speaker was held out for hyperparameter and threshold tuning. 

Speaker-dependent (SD) evaluation provides an empirical upper bound. Each retained speaker's clips were split into five stratified folds preserving that speaker's dysarthric prevalence, and a classifier was trained and evaluated on each speaker independently. Speakers with fewer than ten dysarthric clips were excluded, since stratified five-fold splitting requires at least two positive clips per fold, leaving nine of eleven speakers. Per-speaker scores were averaged. This protocol is clinically meaningful in AC, where a pre-operative recording of the same patient can serve as training data for an intra-operative classifier. The difference between SD and SID therefore quantifies the cost of transferring across speakers rather than calibrating to one. 

The LOSO evaluation holds out a single speaker per fold and repeats across all eleven speakers, and is the strictest form of speaker-independent evaluation. With eleven speakers, LOSO produced cross-fold AUC standard deviations near $0.10$, against approximately $0.04$ under SID grouping. The two protocols agree on point estimates, which indicates that the 5-fold grouping does not bias the reported AUC. 

Because per-speaker clip counts and dysarthric proportions vary widely (Table~\ref{tab:databrase_distribution}), metrics are decomposed by speaker so that individual contributions remain visible rather than absorbed into a cohort average. A single global decision threshold is unsuitable for the same reason, and the operating point is instead computed from each held-out speaker's own NT distribution.

\subsection{Experimental design} 
\label{sec:exp_design} 

\begin{figure}[t]
	\centering
	\begin{tikzpicture}[
		font=\footnotesize,
		stage/.style={
			rectangle, draw, rounded corners=3pt,
			text width=64mm, align=left, inner sep=2mm
		},
		s1/.style={stage, fill=blue!8},
		s2/.style={stage, fill=cyan!10},
		s3/.style={stage, fill=green!8},
		arrow/.style={-{Latex[length=2mm]}, line width=0.7pt, gray!70},
		]
		
		\node[s1] (s1) {
			\textbf{Phase 1 \textperiodcentered\ Baselines}\\[1pt]
			XGBoost, MLP, BiGRU on conventional features
		};
		
		\node[s2, below=4mm of s1] (s2) {
			\textbf{{+} Data augmentation}\\[1pt]
			HiFi-GAN pitch shift, corrects class imbalance
		};
		
		\node[s3, below=4mm of s2] (s3) {
			\textbf{Phase 2 \textperiodcentered\ {+} Cascade models}\\[1pt]
			Cascade-MLP, Cascade-BiGRU
		};
		
		\node[s3, below=4mm of s3] (s4) {
			\textbf{{+} Diarization}\\[1pt]
			Patient-frame extraction, conversational features
		};
		
		\node[s3, below=4mm of s4] (s5) {
			\textbf{{+} Self-supervised features}\\[1pt]
			Multi-layer wav2vec2 $\{6,9,\mathrm{final}\}$ fusion
		};
		
		\draw[arrow] (s1.south) -- (s2.north) node[midway,right,font=\scriptsize,text=gray]{inherits};
		\draw[arrow] (s2.south) -- (s3.north) node[midway,right,font=\scriptsize,text=gray]{inherits};
		\draw[arrow] (s3.south) -- (s4.north) node[midway,right,font=\scriptsize,text=gray]{inherits};
		\draw[arrow] (s4.south) -- (s5.north) node[midway,right,font=\scriptsize,text=gray]{inherits};
		
	\end{tikzpicture}
	\caption{Cumulative experimental design.}
	\label{fig:exp_design}
\end{figure}
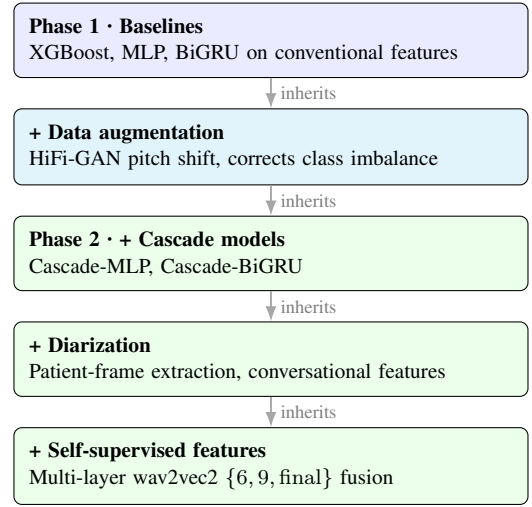

The individual effect of each component cannot be read from the performance of the complete framework. Experiments are therefore organized as an ordered sequence of additions (Fig.~\ref{fig:exp_design}), each introducing one component while the others are held fixed. 

Baseline classifiers on the conventional features establish the reference. Three estimators of different inductive biases are compared, since a component-wise analysis is meaningful only if the choice of estimator does not dominate. Augmentation is examined next under both SD and speaker independent evaluation, which separates its two possible effects, namely a better estimate of a given speaker's dysarthric range or improved transfer to unseen speakers. It is retained thereafter because it addresses the class imbalance of Table~\ref{tab:databrase_distribution}, independently of its effect on either protocol. 

The cascades are introduced with the same features as the baselines, so any difference isolates the two-stage architecture. The remaining experiments act on the representation. Self-supervised extraction from patient frames is compared against full-clip extraction, and the feature families are then added cumulatively, from the cepstral baseline to the multi-layer fusion. Reporting the classifier comparison and the representation additions on the same cohort and protocols allows the two to be weighed against each other.

\section{Results}
\label{sec:results}

The framework was evaluated on the retained cohort of eleven patients, comprising 2{,}706 clips after the quality criterion, with a dysarthric prevalence of 22.1\%. Each component is assessed in the order it is introduced in Section~\ref{sec:exp_design}, so that its effect is measured against the configuration preceding it. The conventional feature set and the three single-stage classifiers establish the reference, augmentation is assessed on that same configuration, the cascades are introduced without altering the representation, and the representation is varied last. AUC is the primary metric throughout, and results are reported under the protocols defined in Section~\ref{sec:cv}.

\subsection{Baseline classifiers and data augmentation}

Three classifiers of distinct inductive bias were first evaluated on the conventional feature set under the LOSO protocol (Table~\ref{tab:baselines1}). XGBoost (0.6148) and the MLP (0.6117) perform comparably, while the BiGRU (0.5873) is weakest, and the leading metric is split across the three systems. The spread of $0.028$ AUC is small relative to the cross-fold variability of this cohort, indicating that architecture is not the limiting factor on this representation. All three remain close to chance-corrected performance, with F1 below $0.38$, which is consistent with a representation that transfers poorly across speakers.

\begin{table}[h]
	\centering
	\caption{Baseline classifier performance under the LOSO protocol.}
	\label{tab:baselines1}
	\begin{tabular}{lcccc}
		\toprule
		Classifier & AUC & F1 & BAcc & Acc \\
		\midrule
		XGBoost & 0.6148 & 0.3782 & 0.6334 & 0.6298 \\
		MLP     & 0.6117 & 0.3793 & 0.5988 & 0.5899 \\
		BiGRU   & 0.5873 & 0.3629 & 0.6203 & 0.6402 \\
		\bottomrule
	\end{tabular}
\end{table}

\begin{figure*}[t]
	\centering
	\includegraphics[width=\textwidth]{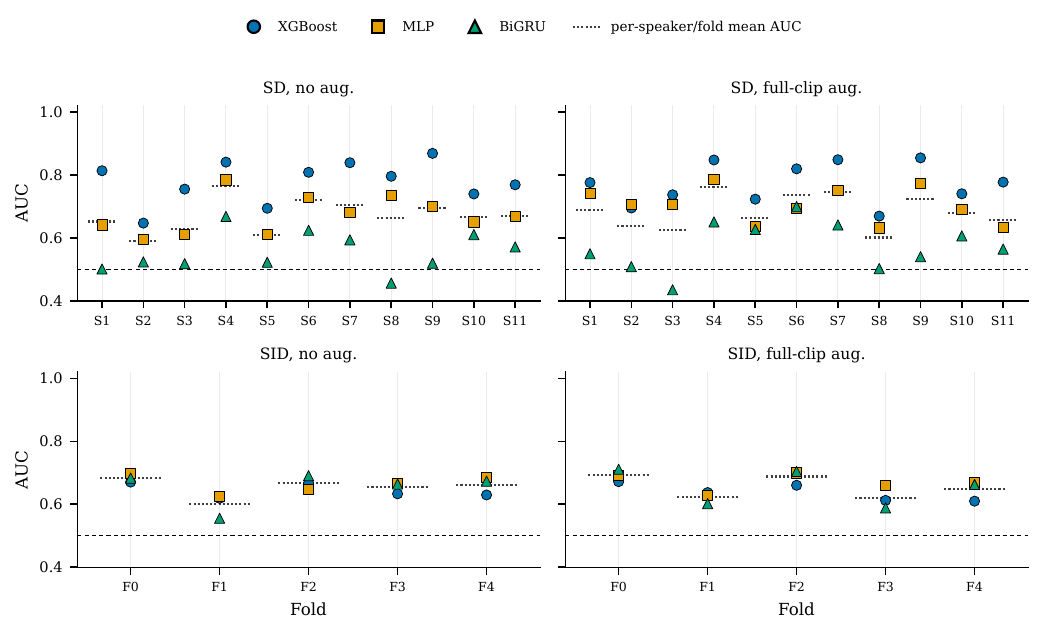}
	\caption{Per-unit AUC for the three baseline classifiers under each protocol and augmentation condition, on the conventional feature set.}
	\label{fig:grid1}
\end{figure*}

The same systems were then assessed with and without full-clip HiFi-GAN augmentation under the SD and SID protocols (Table~\ref{tab:augmentation1}). Augmentation yields no systematic benefit. It improves the MLP under SD (0.6946 against 0.6550) and the BiGRU under the SID protocol (0.6141 against 0.5597), but degrades XGBoost under the SID protocol (0.5872 against 0.6440) and the BiGRU under SD (0.4699 against 0.5823). The direction of the effect is therefore conditional on both the estimator and the evaluation regime, and no system benefits under both protocols. This is consistent with augmentation acting on within-speaker variability, which the SD regime rewards, without supplying the inter-speaker variability that cross-speaker generalization requires. Per-speaker and per-fold distributions are shown in Fig.~\ref{fig:grid1}.

\begin{table*}[t]
	\centering
	\caption{Effect of full-clip data augmentation under the SD and SID protocols.}
	\label{tab:augmentation1}
	\begin{tabular}{llcccccccc}
		\toprule
		& & \multicolumn{4}{c}{SD (per-speaker 5-fold CV)} & \multicolumn{4}{c}{SID (5-fold grouped)} \\
		\cmidrule(lr){3-6}\cmidrule(lr){7-10}
		Classifier & Augmentation & AUC & F1 & BAcc & Acc & AUC & F1 & BAcc & Acc \\
		\midrule
		\multirow{2}{*}{XGBoost}
		& None      & 0.8496 & 0.6054 & 0.7237 & 0.8245 & 0.6440 & \textbf{0.3671}   & 0.6303 & 0.6545 \\
		& Full-clip & 0.8508 & 0.6066 & 0.7079 & 0.8229 & 0.5872 & 0.3592 & 0.6228 & 0.6041 \\
		\midrule
		\multirow{2}{*}{MLP}
		& None      & 0.6550 & 0.3935 & 0.6505 & 0.6901 & 0.6434 & 0.3939 & 0.6373 & 0.6357 \\
		& Full-clip & 0.6946 & 0.4226 & 0.6608 & 0.7163 & 0.6345 & 0.3857 & 0.6261 & 0.6850 \\
		\midrule
		\multirow{2}{*}{BiGRU}
		& None      & 0.5823 & 0.3698 & 0.5357 & 0.4007 & 0.5597 & 0.3450 & 0.6181 & 0.5427 \\
		& Full-clip & 0.4699 & 0.3469 & 0.5518 & 0.3907 & 0.6141 & 0.3823 & 0.6000 & 0.6166 \\
		\bottomrule
	\end{tabular}
	\par\smallskip
\end{table*}

The two protocols differ substantially. XGBoost attains 0.8508 under SD against 0.6440 under the SID protocol, and comparable margins hold for the other systems. This gap between calibrating to a patient and generalizing to an unseen one exceeds any effect attributable to the classifier or to augmentation, and it persists across the experiments that follow.

\subsection{Cascade architectures} 
\label{subsec:cascade} 

The two cascades were evaluated against their single-stage counterparts on the same feature set, under the SID and LOSO protocols (Table~\ref{tab:cascade1}), so that any difference is attributable to the architecture alone. 

\begin{table*}[t]
	\centering
	\caption{Cascade architectures performance under the SID five-fold and LOSO protocols.}
	\label{tab:cascade1}
	\begin{tabular}{llcccc}
		\toprule
		Protocol(SID) & System & AUC & F1 & BAcc & Acc \\
		\midrule
		\multirow{4}{*}{5-fold}
		& MLP    & 0.6434 & 0.3939 & 0.6373 & 0.6357 \\
		& Cascade-MLP          & 0.6428 & 0.3884 & 0.6119 & 0.5563 \\
		& BiGRU & 0.5597 & 0.3450 & 0.6181 & 0.5427 \\
		& Cascade-BiGRU        & 0.5771 & 0.3531 & 0.5640 & 0.5128 \\
		\midrule
		\multirow{4}{*}{LOSO}
		& MLP   & 0.6117 & 0.3793 & 0.5988 & 0.5899 \\
		& Cascade-MLP          & \textbf{0.6669} & 0.4049 & 0.6308 & 0.5698 \\
		& BiGRU  & 0.5873 & 0.3629 & 0.6203 & 0.6402 \\
		& Cascade-BiGRU        & 0.5968 & 0.3653 & 0.5958 & 0.5374 \\
		\bottomrule
	\end{tabular}
\end{table*}

The benefit of the cascade is confined to the stricter protocol. Under LOSO, the Cascade-MLP attains 0.6669 against 0.6117 for the single-stage MLP, and the Cascade-BiGRU attains 0.5968 against 0.5873, so both cascades exceed their baseline. Under the SID protocol the pattern does not hold: the Cascade-BiGRU improves marginally (0.5771 against 0.5597) while the Cascade-MLP does not (0.6428 against 0.6434). 

This dependence on protocol is informative. The two protocols differ in how many speakers are held out per fold, so LOSO trains on more speakers and tests on a single unseen one. The first-stage estimate is generated by an inner LOSO procedure, and its usefulness to the second stage depends on how well it transfers to a speaker the tree has not seen. The cascade therefore helps where the first stage is estimated from the widest speaker sample, and contributes little when the training set is smaller and the test set more heterogeneous. 

The gain is also larger for the feed-forward second stage than the recurrent one. The Cascade-MLP improves on its baseline by a wider margin than the Cascade-BiGRU under LOSO, which is consistent with the difference in how the first-stage estimate is admitted: the MLP receives it as an additional input dimension alongside the features, whereas the BiGRU receives only its session-level mean at the classification head. Even at its best, however, the Cascade-MLP remains near 0.67 AUC under LOSO, well below the 0.85 attained under speaker-dependent evaluation, so the architecture does not close the cross-speaker gap.


\subsection{Feature representation} 
\label{subsec:representation} 

Two experiments vary the representation while the classifiers remain fixed: the region from which the self-supervised block is extracted, and the cumulative addition of feature families. 

\begin{figure}
    \centering
     \includegraphics[width=0.8\linewidth]{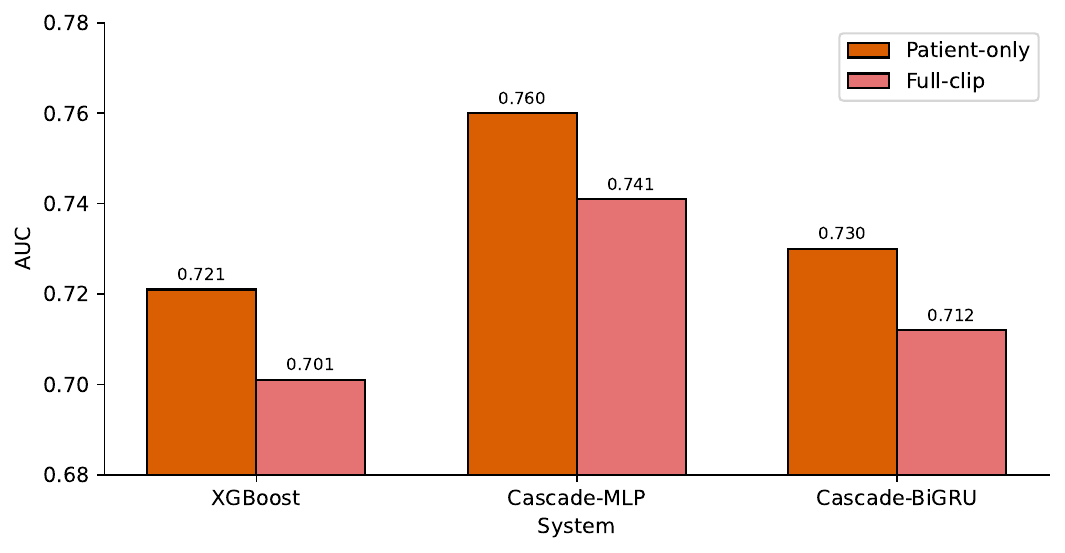}
    \caption{Diarization impact on model performance.}
    \label{fig:Diarization}
\end{figure}

Restricting self-supervised extraction to patient frames improves performance for every system tested (Fig.~\ref{fig:Diarization}). The Cascade-MLP attains 0.760 against 0.741 for full-clip extraction, the Cascade-BiGRU 0.730 against 0.712, and XGBoost 0.721 against 0.701. The improvement is modest but consistent in direction and comparable in magnitude across three architectures of different inductive bias, which indicates that it originates in the representation rather than in any property of a particular estimator. Since the conventional descriptors are unchanged between conditions, the difference is attributable to the removal of clinician speech from the region over which the embedding is computed. 

In the cumulative ablation, AUC increases at each successive configuration for every system (Table~\ref{tab:ablation1}). From the cepstral baseline to the proposed multi-layer fusion, the Cascade-MLP rises from 0.643 to 0.760, XGBoost from 0.591 to 0.721, and the Cascade-BiGRU from 0.579 to 0.730. For all three systems, each configuration improves on the previous one, with the Cascade-MLP highest throughout. The multi-layer fusion outperforms the single-layer configuration for each system, supporting the layer-wise argument that articulatory precision, phonological realization, and intelligibility are encoded at different depths, so that no single layer represents the impairment completely.

Taken together, the two sets of experiments locate the source of performance. Across the ablation, the representation accounts for a gain of between 0.117 and 0.151 AUC depending on the system, whereas the three architectures differ by at most 0.070 at any fixed configuration. The representation therefore governs cross-speaker performance to a greater extent than the choice of classifier, and the cascaded architectures improve on their baselines without altering that conclusion.

\begin{table*}[t]
	\centering
	\caption{Cumulative feature ablation under the SID five-fold protocol.}
	\label{tab:ablation1}
	\begin{tabular}{llcccc}
		\toprule
		Configuration & System & AUC & F1 & BAcc & Acc \\
		\midrule
		\multirow{3}{*}{MFCC only}
		& XGBoost       & 0.591 & 0.381 & 0.545 & 0.313 \\
		& Cascade-MLP   &  {0.643} & 0.407 & 0.601 & 0.486 \\
		& Cascade-BiGRU & 0.579 & 0.371 & 0.563 & 0.455 \\
		\midrule
		\multirow{3}{*}{+ Full Phase~1 (incl. diarization)}
		& XGBoost       & 0.662 & 0.434 & 0.635 & 0.639 \\
		& Cascade-MLP   &  {0.699} & 0.446 & 0.654 & 0.685 \\
		& Cascade-BiGRU & 0.629 & 0.405 & 0.613 & 0.642 \\
		\midrule
		\multirow{3}{*}{+ wav2vec2 layer 6}
		& XGBoost       & 0.701 & 0.453 & 0.651 & 0.700 \\
		& Cascade-MLP   &  {0.737} & 0.472 & 0.678 & 0.691 \\
		& Cascade-BiGRU & 0.664 & 0.429 & 0.635 & 0.648 \\
		\midrule
		\multirow{3}{*}{+ wav2vec2 $\{6,9,\mathrm{final}\}$ (proposed)}
		& XGBoost       & 0.721 & 0.461 & 0.654 & 0.671 \\
		& Cascade-MLP   &  {0.760} & 0.494 & 0.694 & 0.710 \\
		& Cascade-BiGRU & 0.730 & 0.477 & 0.669 & 0.683 \\
		\bottomrule
	\end{tabular}
	\par\smallskip
\end{table*}

\section{Discussion}
\label{sec:discussion}
The stage-wise design summarized in Fig.~\ref{fig:exp_design} attributes the observed performance to individual pipeline components rather than to the system as a whole. The results show that the classifier is a secondary factor, while the successive representational stages (cascading, diarization, and the self-supervised feature set) account for the largest cross-speaker gains.

The baseline comparison (Table~\ref{tab:baselines1}) shows that three classifiers which learn in fundamentally different ways, a gradient-boosted tree, a feed-forward network, and a recurrent network, span a cross-speaker AUC range of only $4.7\%$ under the LOSO protocol, from $0.5873$ (BiGRU) to $0.6148$ (XGBoost). The convergence of such different architectures to within a few percent is hard to reconcile with the classifier being the limiting factor. If the estimator is varied across its full plausible range and performance changes a little, then the ceiling on performance comes from the information present in the input, not from the model's ability to use it. The results support the same conclusion, since all three systems sit near an AUC of $0.6$, far below clinically useful discrimination, which suggests that conventional acoustic descriptors carry only a weak and largely speaker-specific signal that fails to transfer to unseen patients. Raising cross-speaker performance thus depends on improving the representation rather than the classifier, and the remaining pipeline stages should be optimized in turn.

The gap between evaluation protocols quantifies the difficulty these stages confront. Under the speaker-dependent protocol, XGBoost reaches an AUC $31.9\%$ higher than under the speaker-independent protocol ($0.8496$ versus $0.6440$), and the augmentation results (Table~\ref{tab:augmentation1}) show the between-classifier separation present under the SD protocol reducing to a single overlapping band under the SID protocol. The problem is thus one of transfer across speakers, not of separating impaired from normal speech within a speaker. This also explains why full-clip augmentation produced no systematic cross-speaker benefit. Replicating existing clips increases the training set the model already observes but introduces no new inter-speaker variation, which is the axis along which generalization must occur. Augmentation was therefore retained only for correcting class imbalance, and the transfer difficulty is left to the architectural and representational stages that follow.

The first of these stages is the cascade, which introduces an architectural modification to the baseline classifiers rather than a change in features or training data. The single-stage MLP and BiGRU are restructured into two-stage models in which a first-stage XGBoost feeds a calibrated probability into the neural second stage. Evaluated on the conventional feature set, this structural change recovers discriminative structure that a single-stage model of the same family leaves unused. Under the LOSO protocol (Table~\ref{tab:cascade1}), the Cascade-MLP improves on the single-stage MLP by $9.0\%$ in AUC ($0.6117$ to $0.6669$) and the Cascade-BiGRU improves on the single-stage BiGRU by $1.6\%$, with the gain largest under LOSO, the protocol in which every test speaker is unseen and which most closely reflects deployment on a new patient. This gain follows from the modified architecture. The first-stage XGBoost reduces the high-dimensional, largely weakly informative feature vector into a single calibrated probability that is robust to noisy and redundant dimensions, and the restructured second-stage neural model then operates on this smoothed, speaker-agnostic summary alongside the original features, rather than rediscovering the decision boundary from the raw high-dimensional input in a small-sample model where it is prone to overfitting speaker identity. The inner LOSO construction that generates the first-stage probabilities is essential to this validity, since it ensures the second stage is trained on out-of-sample probabilities of the same statistical character it receives at test time, preventing an overconfident in-sample signal that would not transfer.

The next stage acts on the representation itself, through diarization-conditioned extraction. Restricting the self-supervised embedding to diarized patient frames rather than the full clip raises AUC by $2.9\%$ for XGBoost, $2.6\%$ for the Cascade-MLP, and $2.5\%$ for the Cascade-BiGRU (Fig.~\ref{fig:Diarization}). The gain is smaller than that of the cascade, but its uniformity across three classifiers of different architectures is the more informative feature. An effect that follows the change in extraction region regardless of the downstream model is a property of the representation's input rather than of any one estimator, and is therefore attributable to the extraction choice itself. Diarization shows that the clinician speech occupies a class-dependent share of each recording, so a full-clip embedding encodes recording-context structure that correlates with the label without reflecting patient pathology. Restricting extraction to patient frames removes this confound at the representation level, before any classifier is applied, which is why the improvement transfers across all three architectures.

The final and largest stage is the complete self-supervised representation, whose contribution the cumulative ablation (Table~\ref{tab:ablation1}) makes explicit. Across the full pipeline, moving from the MFCC baseline to the proposed multi-layer configuration raises AUC by $22.0\%$ for XGBoost, $18.2\%$ for the Cascade-MLP, and $26.1\%$ for the Cascade-BiGRU, gains an order of magnitude larger than the $4.7\%$ that separates the classifiers. Performance is therefore governed primarily by the representation rather than the classifier, justifying the intensive feature-extraction effort in this work.

The structure of this gain is technically informative. Adding the full conventional Phase~1 stack over MFCC contributes a relative AUC gain of $8.6\%$ to $12.0\%$, but this arises from many descriptors acting together and is partly classifier-dependent. A single self-supervised wav2vec2 layer, by contrast, contributes a further $5.4\%$ to $5.9\%$ from one feature family alone. Per feature family, the self-supervised representation is considerably more discriminative signal than the conventional descriptors. This is expected, since hand-specified descriptors largely encode speaker identity and do not transfer, whereas the W2V2 encoder, pretrained on external corpora, captures phonetic and phonological structure learned independently of this cohort and therefore generalizes to new speakers.

The multi-layer fusion is justified by how these encoders represent speech, not just by empirical gain. Extending the single layer to the $\{6,9,\mathrm{final}\}$ fusion adds a further $2.9\%$ to $3.1\%$ for the tree and feed-forward cascades and $9.9\%$ for the recurrent cascade. Layer-wise analyses of self-supervised speech encoders indicate that phonetic detail is concentrated in mid-network layers, phonological structure in intermediate-deep layers, and intelligibility- and speaker-related information in the final layers. Since dysarthria simultaneously impairs articulatory precision, phonological realization, and intelligibility, no single layer encodes the full profile, and the consistent improvement from fusion across all three classifiers indicates that the selected depths contribute complementary rather than overlapping information.

In practical terms, this hierarchy implies that a deployable monitoring system should be built on clean, patient-attributable measurements and a strong pretrained representation, treating the classifier as a largely interchangeable final component. The achievable performance falls between two scenarios. The cross-speaker results correspond to a previously unseen patient evaluated without calibration, whereas the markedly higher speaker-dependent results, exceeding the cross-speaker values by up to $31.9\%$ in AUC, correspond to the case where a pre-operative recording enables patient-specific calibration. Since AC already involves a structured pre-operative assessment, this calibrated regime is clinically realistic, and the difference between the two regimes indicates how much performance a brief pre-operative baseline could recover.

Several limitations remain. The eleven-speaker cohort is small, so the argument rests on the directional consistency of the orderings across classifiers and configurations rather than on individual point estimates, and a larger, more phenotypically diverse cohort would be needed to separate configurations at finer resolution. The binary dysarthria-versus-no-trouble target also collapses severity levels and does not address whether the dysarthric signature varies with the cortical or subcortical stimulation site; a multi-class extension distinguishing dysarthria, anomia, and paraphasia, annotated separately rather than aggregated, would add clinical granularity beyond that of prior AC work. Finally, generalization beyond DATABRASE is untested. The cross-lingual evaluation of~\cite{maoudjClassificationSpeechArrests2025} showed substantial performance loss under Japanese--French transfer, indicating that language- and center-specific factors interact with the self-supervised representation in ways this cohort cannot probe. Validation on independent AC cohorts at different centers, ideally with matched-protocol re-evaluation, together with fine-tuning rather than freezing the encoder once cohort size permits, are the natural next steps before stronger deployment claims can be made.

\section{Conclusion}
\label{sec:conclusion}
This work approached intra-operative dysarthria detection as a component-wise pipeline-optimization problem, quantifying the contribution of each stage from data quality through feature extraction and classification. The results point to a single conclusion that cross-speaker performance is governed principally by the representation rather than the classifier. Three classifiers of fundamentally different design converge to within $4.7\%$ AUC, whereas enriching the representation from conventional descriptors to a multi-layer self-supervised fusion raises AUC by $18.2\%$ to $26.1\%$, an order of magnitude larger, with diarization-conditioned extraction adding a further consistent gain. The proposed cascaded architecture recovers additional structure a single-stage model leaves unused, most clearly under the strict LOSO protocol that best reflects deployment on a new patient. Together, these results imply that the priority should be clean patient-attributable measurement and strong pretrained representations rather than more complex classifiers for low-resource intra-operative settings. The main limitation is the small speaker cohort, and the most promising directions for further work are the acquisition of a larger and more diverse speaker set and the fine-tuning, rather than freezing of the self-supervised encoder once cohort size makes this safe against overfitting to speaker identity.


\bibliographystyle{IEEEtran}
\bibliography{classification_of_dysarthria_AC}

\end{document}